\pdfoutput=1
\documentclass[11pt]{article}

\usepackage[final]{acl}

\usepackage{times}
\usepackage{latexsym}

\usepackage[T1]{fontenc}
\usepackage[utf8]{inputenc}

\usepackage{microtype}

\usepackage{inconsolata}

\usepackage{graphicx}
\usepackage{amsmath}
\usepackage{amssymb}
\title{Mapping Hymns and Organizing Concepts in the Rigveda: Quantitatively Connecting the Vedic Suktas}

\author{
  \textbf{Venkatesh Bollineni},
  \textbf{Igor Crk},
  \textbf{Eren Gultepe\textsuperscript{*}}
\\
Dept. of Computer Science, Southern Illinois University Edwardsville, USA
\\
  \small{
    \textbf{Correspondence:} *\href{mailto:egultep@siue.edu}{egultep@siue.edu}
  }
}

\begin{document}
\maketitle
\begin{abstract}
Accessing and gaining insight into the Rigveda poses a non-trivial challenge due to its extremely ancient Sanskrit language, poetic structure, and large volume of text. By using NLP techniques, this study identified topics and semantic connections of hymns within the Rigveda that were corroborated by seven well-known groupings of hymns. The 1,028 suktas (hymns) from the modern English translation of the Rigveda by Jamison and Brereton were preprocessed and sukta-level embeddings were obtained using, i) a novel adaptation of LSA, presented herein, ii) SBERT, and iii) Doc2Vec embeddings. Following an UMAP dimension reduction of the vectors, the network of suktas was formed using k-nearest neighbours. Then, community detection of topics in the sukta networks was performed with the Louvain, Leiden, and label propagation methods, whose statistical significance of the formed topics were determined using an appropriate null distribution. Only the novel adaptation of LSA using the Leiden method, had detected sukta topic networks that were significant (\textit{z} = 2.726, \textit{p} < .01) with a modularity score of 0.944. Of the seven famous sukta groupings analyzed (e.g., creation, funeral, water, etc.) the LSA derived network was successful in all seven cases, while Doc2Vec was not significant and failed to detect the relevant suktas. SBERT detected four of the famous suktas as separate groups, but mistakenly combined three of them into a single mixed group. Also, the SBERT network was not statistically significant. 

\end{abstract}

\section{Background and Significance}

The Rigveda is written in Vedic Sanskrit and is the oldest existing sample of Sanskrit literature, written approximately 3000 years ago, in the region of present-day Afghanistan and the Punjab region of India \citep{jamison2014rigveda}. It is a heterogeneous collection of hymns (suktas) written by various poets (Rishis), that praise gods, describe rituals, and provide wisdom \citep{jamison2014rigveda, rigvedaOrganization}. Popular mantras recited by Hindus, such as the Gayatri mantra, is chanted at three different times of the day \citep{smith2020invisible} for the purposes of mental well-being, and the Mahamrityunjaya mantra, which is recited for physical protection and longevity, are both sourced from the Rigveda \citep{devananda1999meditation}. 

Yet despite being a central text in Hinduism, navigating the Rigveda and obtaining insights regarding concepts and topics are not as straightforward as the Bible or Quran, for which there are innumerable resources (such as commentaries) and written for individuals at varying levels of skill and familiarity with the books. This is especially true for individuals who do not speak or understand any of the Indian languages. Although scholarly articles regarding specific topics (such as death) in the Rigveda are available, for the layperson interested in learning about the Rigveda, organizing and collating the information may be unwieldy \citep{jamison2014rigveda}. This is further evidenced in NLP studies, where the quantity of studies focused on the Abrahamic religions vastly outnumbers those focused on Hindu religious texts \citep{hutchinson-2024-modeling}.

\section{Related Work}

Recent studies have analyzed Hindu religious and literary texts from various aspects. One study extracted and formed social networks among the Pandavas (protagonists) and Kuaravas (antagonists) in the Mahabharata (an epic poem from the Hindu scriptures) using matrix factorization and spectral graph theory techniques \citep{gultepe2023}. In another study, using linguistic and lexical features in Sanskrit, the Mahabharata was stratified into clusters \citep{StratifyingtheMahbhrata}. Another study had determined topics on the English translations of two other important Hindu texts \citep{chandra2022artificial}, the Upanishads and the Bhagavad Gita, using pre-trained sentence embeddings obtained from deep learning networks, Sentence Embeddings using Siamese BERT-Networks (SBERT) \citep{reimers-2019-sentence-bert} and Universal Sentence Encoder (USE) embeddings \citep{cer-etal-2018-universal}. Many hymns in the Rigveda can be attributed to specific devas (deities in Hinduism), such as \textit{Indra} and \textit{Agni} and were predicted using neural network-based word embedding models such as Word2Vec \citep{mikolov2013distributed} and GloVe \citep{pennington-etal-2014-glove} with linear classifiers \citep{mahesh-bhattacharya-2023-creation}.

Many studies have focused on modelling the syntactics and parsing of the Sanskrit language using various deep learning techniques, such as recurrent neural networks \citep{aralikatte-etal-2018-sanskrit,hellwig-nehrdich-2018-sanskrit} or transformers \citep{sandhan-etal-2022-translist,hellwig_data-driven_2023,nehrdich-etal-2024-one} to generate new Sanskrit text. Another use has been to create sentence and word embeddings using transformers and static models for semantic and analogy tasks \citep{lugli-etal-2022-embeddings}. Some studies have shown that combining semantic information from the Sanskrit Word Net \citep{Short_Luraghi_Biagetti} with parsed sentences in the Vedic TreeBank \citep{hellwig-etal-2020-treebank} can help to provide better understanding of sentence structure \citep{biagetti-2023-linking} and may improve Sanskrit language modelling using Sanskrit neural word embeddings \citep{sandhan-etal-2023-evaluating}

Only a handful of studies have focused on clustering or stratifying Vedic texts for the purposes of obtaining insights about texts written in Vedic Sanskrit. One such study had performed Bayesian mixture modelling to obtain a chronological ordering of texts written in Vedic Sanksrit, such as the Rigveda, Atharvaveda, and post-Rigvedic Sanskrit, such as the Aitareya Brahmana \citep{hellwig-2020-dating}. A similar study had analyzed the similarity of passages within the Maitrayani and Kathaka Samhitas using word embeddings \citep{miyagawa-etal-2024-exploring}. Another study had performed clustering on the linguistic and textual features of the 10 books in the Rigveda to determine whether the historical order of these books can be obtained in a data-centric way \citep{hellwig2021RigStrata}. This study showed that the stratification generally followed the historical divisions. 

The Rigveda has been historically divided into ten books of which the oldest parts are Books II to VII (called the “Family Books”), followed by Books I, VIII, and IX which are accepted to be younger than the Family Books, and Book X is the youngest \citep{jamison2014rigveda}. However, no study has directly investigated the possible organization of the suktas in the Rigveda using NLP techniques such as word, sentence, or document embeddings.

\section{Aim and Contribution}

Thus, the goal of this study was to organize the network of related suktas and uncover the topics contained within the 10 books of the Rigveda in a data-driven manner, without employing prior knowledge about the suktas or topics. Potentially providing a guide for individuals unfamiliar with this complex and varied religious text. This endeavour was mainly facilitated by a novel innovation presented in this study, which we call mean-LSA, where the document vectors obtained using LSA (latent semantic analysis) \citep{deerwester_indexing_1990} were computed from the original length of each sukta (document). This was accomplished by taking the average of all LSA word vectors in a sukta. This is in contrast to obtaining the sukta vectors from suktas that were split into a pre-specified document length, which generally causes a loss of semantic information in normal LSA document vectors. 

Another innovation of this study was that the significance of the sukta networks and detected topics were assessed using a null distribution formed by a random permutation of the network adjacency matrices. Although network structure and topics detected may appear well clustered and organized, i.e. visually the documents appear to be clustered with clear structure, the structure may be due to chance occurrence or an inducement of the preprocessing. This test provides an unbiased method of assessing whether real network structure has been found. Using both innovations, this study demonstrated that historically relevant topics in the Rigveda were detected using the mean-LSA embeddings and were more significant and accurate than those obtained by using the deep learning embedding techniques of SBERT and Doc2Vec \citep{refMikolovDoc2Vec}, both of which provided non-significant network structure.    

\begin{figure*}[!ht]
\centering
 \includegraphics[width=1\textwidth]{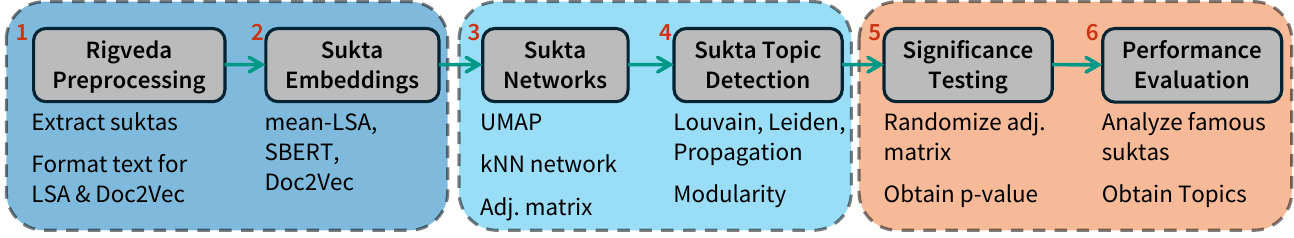} 
	\caption{Processing pipeline for obtaining the network of suktas and topics using the three types of embedding techniques (mean-LSA, SBERT, Doc2Vec). Steps (1) and (2) created the embeddings to form the sukta networks. In steps (3) and (4), using the 4-nearest neighbours of each sukta, the network of topics were detected using community detection methods. Finally, in steps (5) and (6), the statistical significance of the detected network structures were determined and the grouped suktas were analyzed.}
	\label{fig1_pipeline}
\end{figure*}

\section{Methods}

The six steps to obtain the network of hymns (suktas) and topics within the Rigveda is summarized in Figure \ref{fig1_pipeline}. The processing pipeline contained six steps, (1) Rigveda preprocessing to obtain suktas, (2) creation of the sukta embeddings, (3) formation of the sukta similarity network, (4) detection of the topics within the sukta networks, (5) testing of the statistical significance of the sukta networks, and (6) determining the relevance of detected sukta topics. 

\subsection{Rigveda Prepocessing}

To form the network of suktas and detect topics within the Rigveda using word (LSA), sentence (SBERT), or document (Doc2Vec) embeddings, the modern English translation by Jamison and Brereton was used as the source text \citep{jamison2014rigveda}. The Rigveda consists of 10 books (mandalas), 1,028 hymns (suktas), and 10,552 verses (mantras) of varying lengths (Table \ref{tab1_RigOrg}) \citep{jamison2014rigveda,rigvedaOrganization}. Each sukta in the Rigveda is referred to by its mandala and sukta number, e.g., RV 10.129 represents 129\textsuperscript{th} sukta in the 10\textsuperscript{th} mandala, which is the famous Nasadiya sukta in the Rigveda \citep{jamison2014rigveda}.  

\begin{table}[!hb]
%\fontsize{9pt}{9pt}\selectfont
  \centering
  \begin{tabular}{lll}
    \hline
    \textbf{Book} & \textbf{Hymns} & \textbf{Verses}\\
    \hline
    1 & 191 & 2,006\\
    2 & 43 & 429\\
    3 & 62 & 617\\
    4 & 58 & 589\\
    5 & 87 & 727\\
    6 & 75 & 765\\
    7 & 104 & 841\\
    8 & 103 & 1,716\\
    9 & 114 & 1,108\\
    10 & 191 & 1,754\\
    \hline
  \end{tabular}
  \caption{Organization of the documents contained in the Rigveda. Each book (mandala) consists of a collection hymns (suktas), and each hymn is composed of a series of verses (mantras) of varying lengths.}
  \label{tab1_RigOrg}
\end{table}

The three embeddings (LSA, SBERT, Doc2Vec) require slightly different types of text preprocessing. Common to all methods, the text from the Rigveda was organized at the sukta level, in which all the mantras within a sukta were concatenated together and consider as a single document. For LSA, punctuation, numerals, symbols, and stopwords were removed, followed by a conversion to lowercase letters. For the Doc2Vec, a simple preprocessing of converting all uppercase letters to lowercase and tokenization by space was performed \citep{refMikolovDoc2Vec, rehurek2011gensim}. For SBERT, no additional preprocessing was performed \citep{reimers-2019-sentence-bert}.  

\subsection{Sukta Embeddings}

The analysis of the suktas depends on the formation of "sukta embeddings", which are either composed of word, sentence, or document embeddings. In the next subsections, the processing of each method is provided. 

\subsubsection{mean-LSA}

LSA is a classical technique in NLP for obtaining word and document embeddings. Although newer techniques based on deep learning models have been developed, LSA is competitive with methods such as Word2Vec and GloVe on some semantic tasks \citep{levy2015improving}. LSA embeddings are computed using singular value decomposition (SVD) \citep{deerwester_indexing_1990} on unigram and TFIDF weighted data of the suktas, which is represented as $\mathbf{X} \in \mathbb{R}^{v \times n}$, 
where $v$ is the size of the vocabulary, $n$ is the number of suktas, and $d$ is the top singular values (i.e., the dimensionality of the embeddings), giving

\begin{equation}\mathbf{X}_d = \mathbf{U}_d \mathbf{S}_d \mathbf{V}_d^T.\end{equation}
Then, the traditional LSA word embeddings are defined as the rows of 
\begin{equation} \mathbf{W} = \mathbf{U}_d \mathbf{S}_d\end{equation}
and document embeddings are defined as the rows of
\begin{equation} \mathbf{D} = \mathbf{V}_d \mathbf{S}_d.\end{equation}
To obtain both type of LSA embeddings, the suktas must be chunked into equal sized documents. This method will provide unique word embeddings, however, the document embeddings will not represent the original suktas, due to the chunking of the texts. To overcome this hurdle, we introduce an innovation of LSA, where for each sukta, the mean of all the word embeddings $\mathbf{w}_i\in\mathbf{W}$ within the sukta is taken to form the sukta embedding $\mathbf{d}_j^{\textrm{sukta}}$. This method called mean-LSA, creates a unique embedding for each sukta that is representative of the original word length of the sukta.   The mean-LSA embeddings have a dimension of 768, to match the pre-trained SBERT embeddings. 
 
\subsubsection{SBERT}

To obtain the sukta embeddings using SBERT, the
pre-trained 768-dimensional sentence embeddings from the all-mpnet-base-v2 sentence transformer model was used \citep{reimers-2019-sentence-bert}. These embeddings have been trained on 1 billion sentence pairs using the self-supervised contrastive learning objective and is ideal for clustering and  similarity tasks involving sentences and short paragraphs \citep{reimers-2019-sentence-bert}, similar to the length of suktas. The SBERT model is able to handle variable length documents, without any further processing.   

\subsubsection{Doc2Vec}

Doc2Vec ~\citep{refMikolovDoc2Vec}, creates document embeddings that capture semantic and syntactic properties of variable-length documents. A random document embedding is initialized and fine-tuned by predicting words taken from samples in the document. There are two methods for training Doc2Vec, Distributed Memory (DM) and Distributed Bag of Words (DBOW). The DM method concatenates the document embeddings with the word embeddings, to predict the next word in the document. DBOW uses only the document embedding to predict words within the document. The Gensim implementation of Doc2Vec~\citep{refGensim} was used to create 768-dimensional sukta embeddings (to match SBERT) with DBOW and trained for 200 epochs. 

\subsection{Formation of Sukta Networks} 

The sukta embeddings obtained from mean-LSA, SBERT, and Doc2Vec were reduced in dimensionality using Uniform Manifold Approximation and Projection (UMAP) \citep{mcinnes2018umap} to improve computational speed and uncover latent structure among the suktas. Then for each embedding method, the 4 k-Nearest Neighbours (kNN) for each sukta embedding was computed and formed into a binarized adjacency matrix. To determine the nearest neighbours, each sukta embedding was normalized to unit norm and the Euclidean distance was computed. The ranking obtained by the Euclidean distance is identical to that obtained by the cosine distance between the embeddings, although the magnitude of the distances may be different. This procedure creates a network of suktas for each of the embedding methods that captures and summarizes the relationships among the suktas.

\subsection{Community Detection of Topics}

To the detect the community structure within the sukta networks, which may indicate concepts of topics found within the Rigveda, the Louvain \citep{refLouvain}, Leiden \citep{traag2019louvain}, and label propagation algorithms \citep{raghavan2007near} were implemented. The Louvain and Leiden methods attempt to maximize modularity in order to detect communities. Modularity measures the quality of partitioning  a network into communities and ranges from [-1,1] ~\citep{refNewman04} as $Q = \sum_{i}^{}(e_{ii}-a_{i}^2)$, where $e_{ii}$ is the fraction of edges with both nodes in community $i$, and $a_{i}$ is the fraction of edges that attach to nodes in community $i$. The label propagation method attempts to distribute community labels  within a detected community in a semi-supervised manner \citep{raghavan2007near}.

\subsection{Statistical Significance of Topics}

It is necessary to compute the statistical significance of the detected communities within a network to ensure that the observed network structure is not due to chance and the observed groupings represent genuine relationships among the data \citep{kimes_statistical_2017,10.1371/journal.pone.0018961,GULTEPE201813,informatics10040076}. If a high modularity score is obtained, yet with a slight manipulation of the network edges, a similarly high modularity score can be obtained again, then the original modularity score is likely due to a random occurrence of the data. To determine the statistical significance of the network structure, for a predetermined number of iterations, the null distribution was created by randomly permuting the adjacency matrix and performing the relevant network detection method \citep{GULTEPE201813}. This procedure was repeated for 5000 iterations and \textit{p}-value of the original modularity score was obtained by computing the empirical cumulative distribution function \citep{GULTEPE201813}. For all tests the significance level was 5\%.

\subsection{Evaluation of Sukta Topics}

For each embedding method, the topic detection method providing the highest modularity score was chosen and then the significance test was performed. After this two-step procedure, to confirm that the relevant groupings of suktas were obtained, the selected suktas by each network was compared to seven famous grouping of suktas. These sukta groupings were: the Creation, Funeral, and Heaven \& Earth \citep{doniger1981rig}; Marut \citep{muller1869rig}; Surya and Brihaspati \citep{suktapath}, and Water \citep{waterthesis}. 

\subsection{Experimental Setup}

The sukta embeddings from all three methods were row normalized, as it is known to improve representative accuracy \citep{levy2015improving}. After grid search, for UMAP it was found that the best parameters for mean-LSA were: number of neighbours = 8, number of dimensions = 10, min distance = 0.0, and metric = Euclidean. For SBERT, the best UMAP parameters were: number of neighbours = 10, number of dimensions = 5, min distance = 0.0, and metric = Euclidean. For Doc2Vec, the best UMAP parameters were: number of neighbours = 10, number of dimensions = 12, min distance = 0.0, and metric = Euclidean. The best network topic detection for mean-LSA, SBERT, and Doc2Vec were obtained Leiden with Dugue modularity, Louvain with Newan modularity, and Louvain with Potts modularity, respectively. 

\section{Results}

\begin{figure}[!h]
\centering
 \includegraphics[width=1\columnwidth]{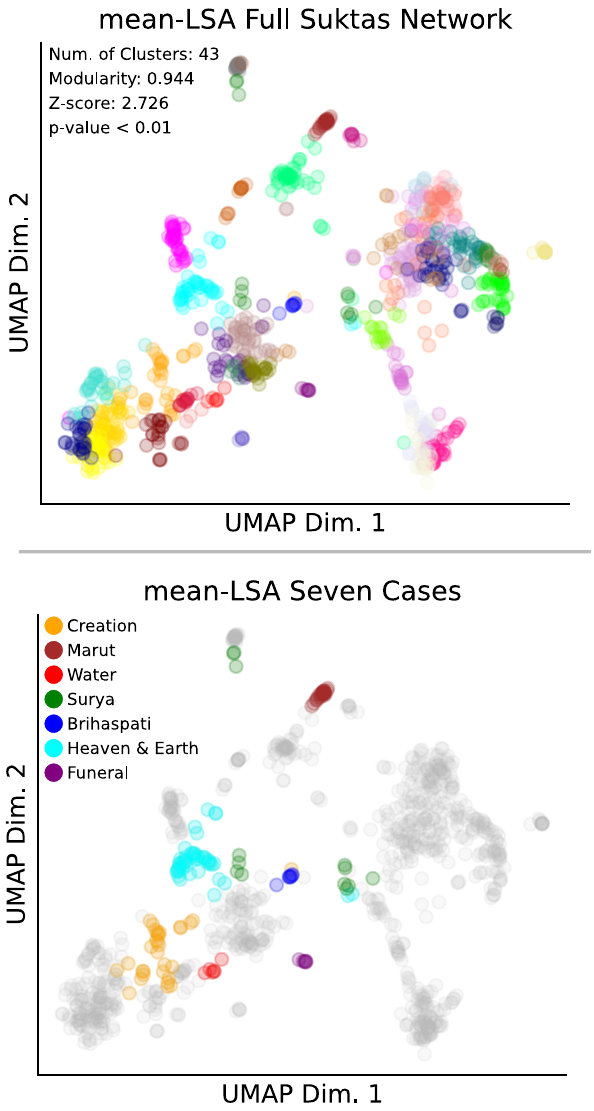} 
	\caption{UMAP visualization of the Rigveda sukta network derived from mean-LSA embeddings. Top: The full network representation, shows 43 unique clusters with a modularity of 0.944 that has statistically significance structure (\textit{z} = 2.726, \textit{p} < .01). Bottom: The highlighted clusters represent a subset of seven famous sukta topics  - Creation, Marut, Water, Surya, Brihaspati, Heaven \& Earth, and Funeral. The mean-LSA embedding network was successful in identifying clusters that contained the semantically related suktas in all seven cases.}
	\label{fig2_meanlsa}
\end{figure}

\begin{figure}[!h]
\centering
 \includegraphics[width=1\columnwidth]{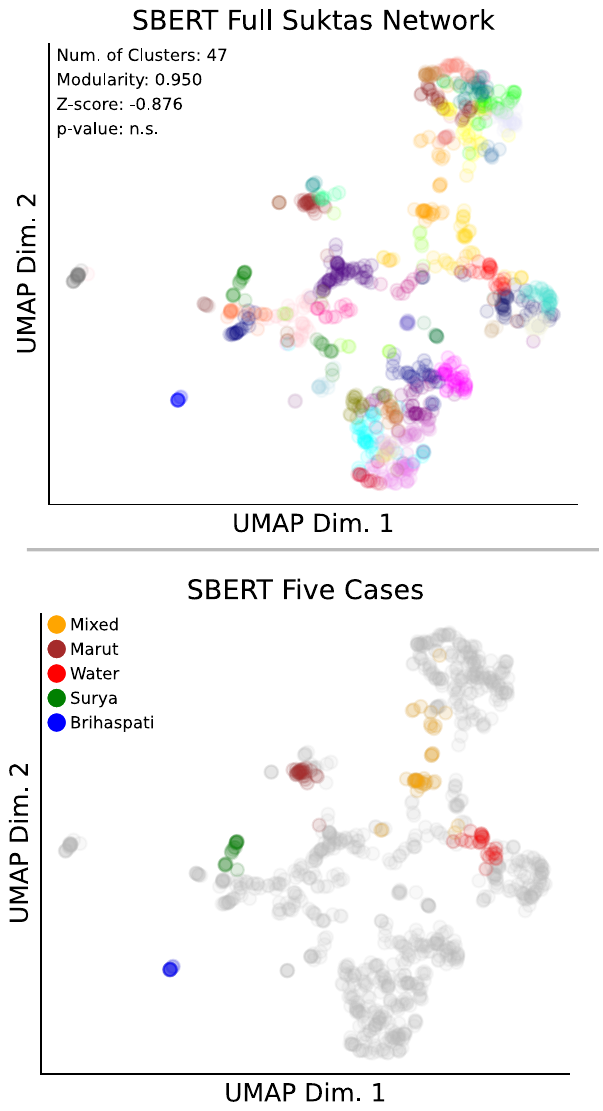} 
	\caption{UMAP visualization of the Rigveda sukta network derived from SBERT embeddings. Top: The full network representation, shows 47 distinct clusters with a modularity of 0.950. Although SBERT's modularity is slightly higher than mean-LSA's modularity (0.944), it failed the significance test (\textit{z} = -0.876, \textit{p} = .810). Bottom: SBERT failed to separate three different topics of suktas (Creation, Funeral, Heaven \& Earth suktas) and clustered them into a single cluster (Mixed). }
	\label{fig3_sbert}
\end{figure}

\begin{figure}[!h]
\centering
 \includegraphics[width=0.99\columnwidth]{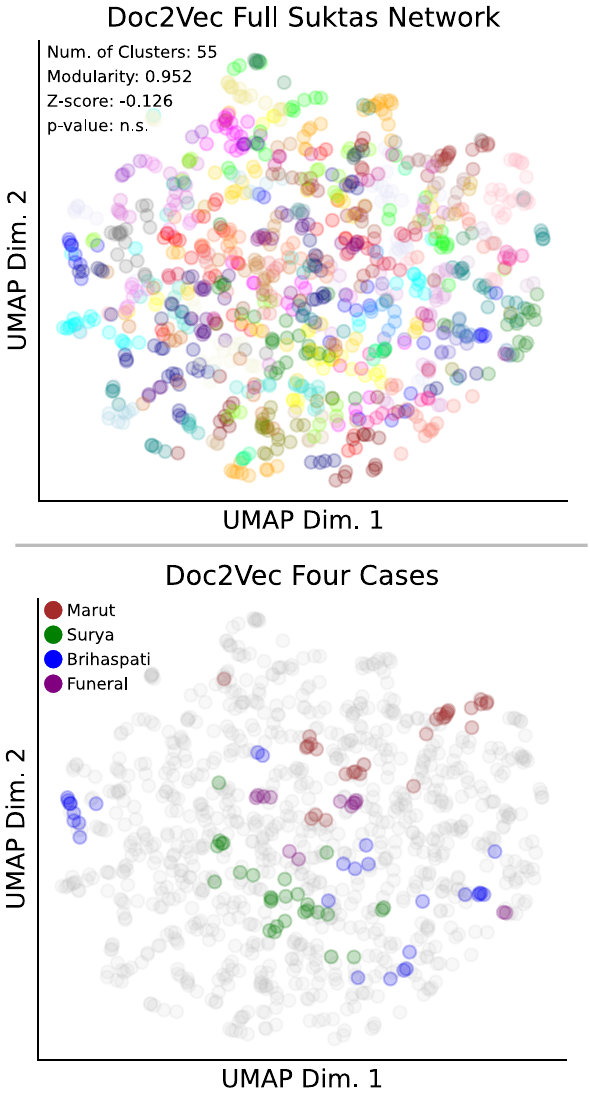} 
	\caption{UMAP visualization of the Rigveda sukta network derived from Doc2Vec embeddings. Top: The full network depicts 55 individual clusters with modularity of 0.952, which is the highest among the three sukta embeddings methods. Despite having higher modularity, it was unsuccessful in passing the statistical significance test (\textit{z} = -0.126, \textit{p} = .550). Bottom: For three out of the seven famous cases, Doc2Vec failed to group the semantically related suktas into relevant clusters and for the four remaining cases (Marut, Surya, Brihaspati, Funeral) the suktas were irregularly distributed.}
	\label{fig4_doc2vec}
\end{figure}

\begin{figure}[!h]
\centering
 \includegraphics[width=1\columnwidth]{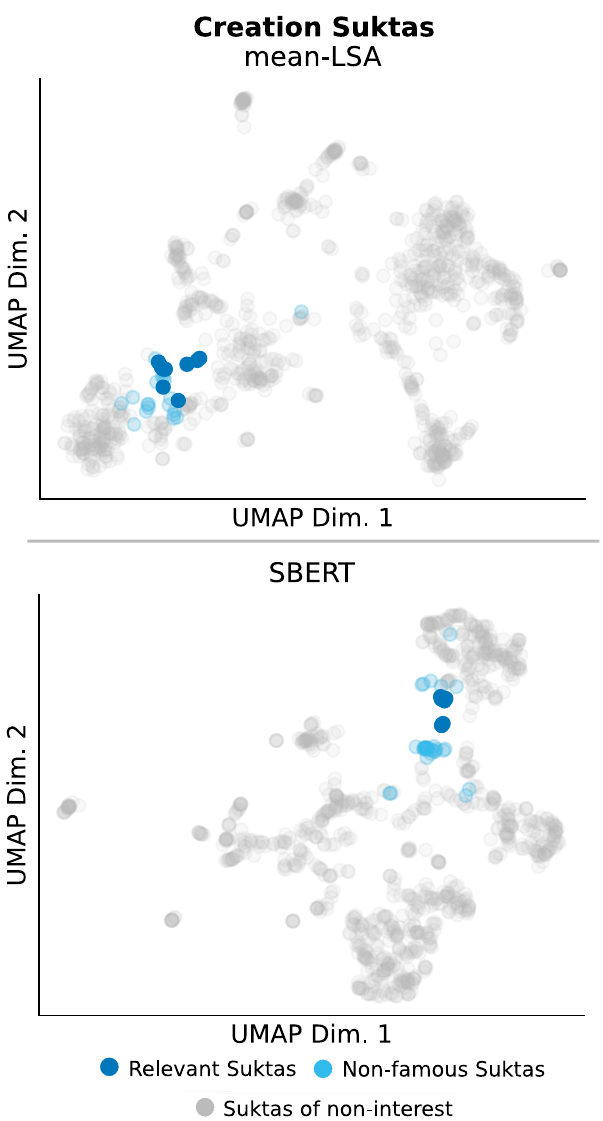} 
	\caption{Comparison of the Creation sukta clusters for the mean-LSA and SBERT sukta embeddings. Top: The network of famous Creation suktas using mean-LSA has gathered all the well-known nine suktas (relevant suktas) into a single cluster with 22 other non-famous suktas. Bottom: SBERT has categorized eight of the nine popular creation suktas together. However, this cluster also contains suktas from other two topics (Funeral and Heaven \& Earth), indicating that SBERT failed to distinguish suktas belonging to other topics.}
	\label{fig5_creation}
\end{figure}

\begin{figure}[!h]
\centering
 \includegraphics[width=0.99\columnwidth]{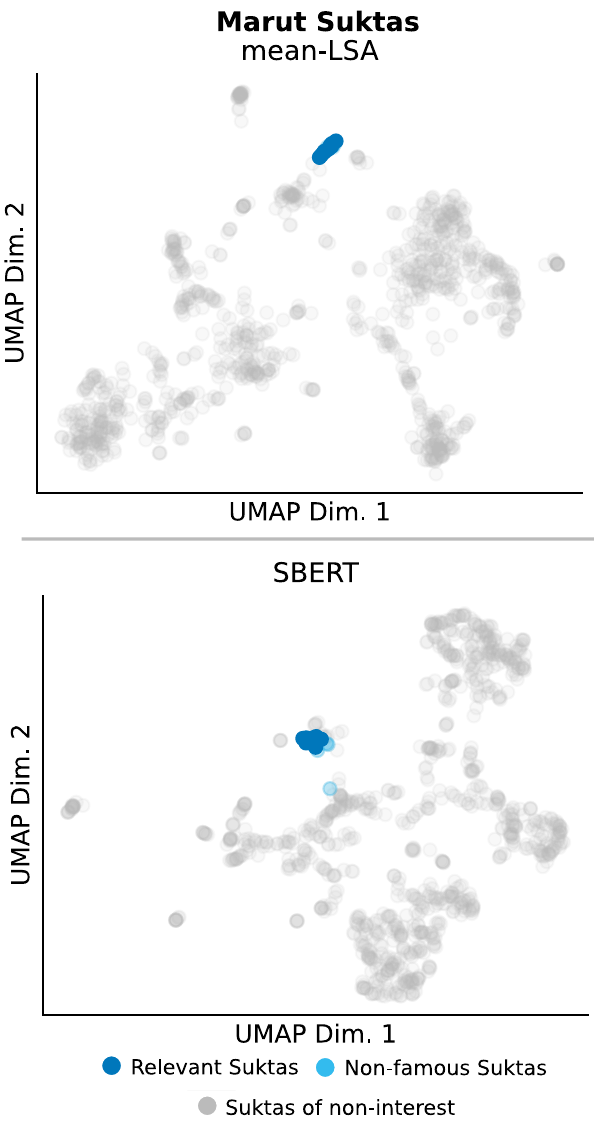} 
		\caption{Comparison of the Marut sukta clusters for the mean-LSA and SBERT sukta embeddings. Top: mean-LSA has clustered ten relevant Marut suktas out of the total 14 famous suktas, alongside 14 other non-famous suktas. Bottom: In the case of SBERT, 12 out of the 14 famous Marut suktas, only two were missing and were placed together with 15 non-famous suktas.}
	\label{fig6_marut}
\end{figure}

\begin{figure}[!h]
\centering
 \includegraphics[width=0.99\columnwidth]{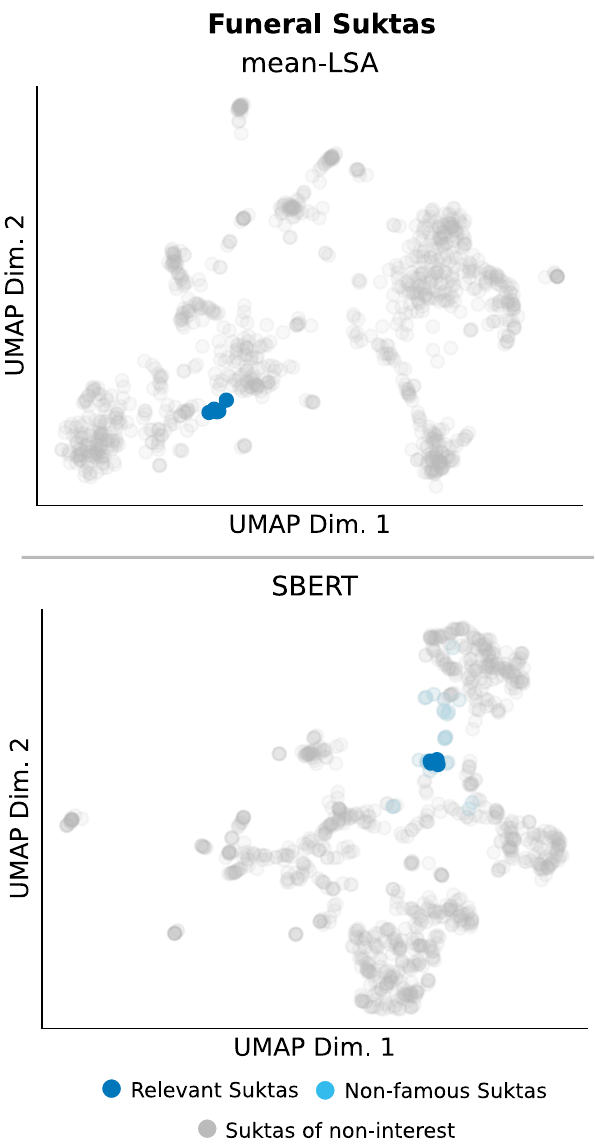} 
	\caption{Comparison of the Funeral sukta clusters for the mean-LSA and SBERT sukta embeddings. Top: mean-LSA successfully captured four out of the six famous funeral suktas along with two Yama (God of Death) suktas, which are also related to funerals. With a total cluster size of six suktas, mean-LSA only identified suktas related to funerals and Yama, without including any non-famous suktas. Bottom: SBERT clustered four suktas related to funerals, consisting of one famous funeral sukta along with three Yama suktas. It mistakenly also captured four suktas related to other topics (Creation, and Heaven \& Earth), indicating that SBERT struggled to separate the suktas based on their topics.}
	\label{fig7_funeral}
\end{figure}

Figures \ref{fig2_meanlsa} (mean-LSA), \ref{fig3_sbert} (SBERT), and \ref{fig4_doc2vec} (Doc2Vec) show all the detected topic clusters and the grouping of the famous clusters obtained by each of the three sukta embedding methods. The mean-LSA sukta embedding method obtained the best sukta organization, as it was the only significant method (\textit{z} = 2.726, \textit{p} < .01) and was successful in identifying clusters that contained the semantically related suktas for all seven cases. Figures \ref{fig5_creation}, \ref{fig6_marut}, and \ref{fig7_funeral} demonstrate how well the mean-LSA sukta embeddings detected the relevant suktas for each case, as compared to SBERT. 

Although, the network of suktas found by SBERT embeddings was not statistically significant (\textit{z} = -0.876, \textit{p} = .810), we still investigated the individual seven famous cases to determine if there were any relevant groupings of the suktas. Overall, the mean-LSA sukta embeddings selected more of the famous suktas at rate of 71.9\% (Table \ref{tab2_meanLSA}) as opposed to the SBERT sukta embeddings which selected the famous suktas at rate of 62.7\% (Table \ref{tab3_sbert}). We did not investigate the Doc2Vec results any further because not only was the network not significant (\textit{z} = -0.126, \textit{p} = .550), there were no meaningful clusters of suktas. 

\begin{table}[!h]
%\fontsize{9pt}{9pt}\selectfont
  \centering
  \begin{tabular}{llll}
    \hline
    \textbf{Case} & \textbf{Correct} & \textbf{Missing}& \textbf{Non-famous}\\
    \hline
    Creation & 9 & 0 & 22\\
    Marut & 10 & 4 & 14\\
    Water & 4 & 2 & 2\\
    Surya & 6 & 10 & 7\\
    Brihaspati & 7 & 2 & 0\\
    H\&E & 6 & 0 & 41 \\
    Funeral & 6 & 6 & 0\\
    \hline
  \end{tabular}
  \caption{Correctly identified famous suktas with mean-LSA. The count of missing famous suktas is also shown along with the selected non-famous suktas. H\&E: Heaven and Earth}
  \label{tab2_meanLSA}
\end{table}

\begin{table}[!h]
%\fontsize{9pt}{9pt}\selectfont
  \centering
  \begin{tabular}{llll}
    \hline
    \textbf{Case} & \textbf{Correct} & \textbf{Missing}& \textbf{Non-famous}\\
    \hline
    Creation & 8 & 1 & 30\\
    Marut & 12 & 2 & 15\\
    Water & 4 & 2 & 21\\
    Surya & 5 & 11 & 12\\
    Brihaspati & 3 & 6 & 7\\
    H\&E & 6 & 0 & 32 \\
    Funeral & 4 & 8 & 34\\
    \hline
  \end{tabular}
  \caption{Correctly identified famous suktas with SBERT. The count of missing famous suktas is also shown along with the selected non-famous suktas. H\&E: Heaven and Earth}
  \label{tab3_sbert}
\end{table}

\section{Discussion and Conclusion}
To our knowledge, this is the first study to create a network of suktas contained in the Rigveda. This is accomplished by using the novel method of mean-LSA, which we presented herein. The mean-LSA method creates a document embedding using the word embeddings obtained from LSA by taking the average of the word embeddings for all words contained in a document. Also, we demonstrated that despite having a high modularity score, this may not be indicative of actual topics found by the network structure. This was corroborated by obtaining the significance values of the network structure through randomization of the network adjacency matrices.

This was further demonstrated by the discrepancy of the modularity scores and the signifance values. The Doc2Vec based sukta network, which had the highest modularity score, did not have a statistically significant structure and it failed to detected any meaningful sukta topic communities, especially in the case of the seven famous suktas. The SBERT based network had a similar situation, in which the modularity score was the second highest, yet it was also not statistically significant. When analyzing the seven famous suktas, it mistakenly combined the Funeral suktas with the Creation, and Heaven \& Earth suktas.     

It may be possible to use the presented statistical significance testing method as a way of determining the cohesiveness and unity of the detected topics. This could be similar to the computation of coherence measures that indicate the relevance of topics against the co-occurrence of words in a topic \citep{roder2015}. However, the statistical test performed with the random permutation of the adjacency matrix may be considering higher-order concepts, since it is manipulating the connections between documents, rather than only analyzing the collection of words. The underlying premise here is that documents are not simply a collection of words. We plan to investigate this application of statistical significance testing of detected topics in a future study. We also plan to investigate the training of unsupervised transformer language models. 

\section{Limitations}
Despite its reliability, the main limitation of this work is that the network analyses relied on a single modern English translation. Thus, as with all translations, the original meaning of the Rigveda in the Vedic Sanskrit may have been masked, since the ability to transmit the true meaning will depend on the ability of the translators to translate the text. For future studies, comparison with the Sanskrit version of the Rigveda is planned. 

\section{Ethics Statement}
The Rigveda is a sacred text in Hinduism and we have been careful to present it in the best way possible, by highlighting important suktas that may be of interest to a wide audience of individuals who may want to learn more about the Hindu religion.

\bibliography{custom}

\end{document}